\documentclass[a4paper, authoryear]{cas-dc}
\usepackage{float}
\usepackage{caption}
\usepackage{algorithmic}
\usepackage{algorithm}
\usepackage[numbers]{natbib}
\def\tsc#1{\csdef{#1}{\textsc{\lowercase{#1}}\xspace}}
\tsc{WGM}
\tsc{QE}
\begin{document}\let\printorcid\relax
\let\WriteBookmarks\relax
\def\floatpagepagefraction{1}
\def\textpagefraction{.001}

\shorttitle{Adaptive Equilibrium: Dynamic Weighting Framework for Generalized Interruption of DeepFake Models}    
\shortauthors{Hongrui Zheng et al}  
%————————————————————————————————————————
% % Main title of the paper
\title [mode = title]{Adaptive Equilibrium: Dynamic Weighting Framework for Generalized Interruption of DeepFake Models}  



\author[1]{Hongrui Zheng}[orcid=0009-0004-1754-3965]
% \cormark[1]
% \ead{107552301310@stu.xju.edu.cn}
% % URL of the first author
% \ead[url]{107552301310@stu.xju.edu.cn}
\affiliation[1]{organization={School of Computer Science and Technology},
            addressline={Xinjiang University}, 
            city={Urumqi},
            country={China}}

\author[1,2,3]{Liejun Wang}[orcid=0000-0003-0210-2273]
\cormark[1]
\ead{wljxju@xju.edu.cn,}% Email id of the second author
% \ead[url]{wljxju@xju.edu.cn,}% URL of the second author
\affiliation[2]{organization={Xinjiang Multimodal Intelligent Processing and Information Security Engineering Technology Research Center},
            city={Urumqi},
            country={China}}
\cortext[1]{Corresponding author}

\author[1,2,3]{Zhiqing Guo}[orcid=0000-0001-6412-334X]
\cormark[1]
\ead{guozhiqing@xju.edu.cn}
% \ead[url]{guozhiqing@xju.edu.cn}
\affiliation[3]{organization={Silk Road Multilingual Cognitive Computing International Cooperation Joint Laboratory},
            addressline={Xinjiang University}, 
            city={Urumqi},
            country={China}}

\begin{abstract}
The advancement of generalized deepfake disruption is constrained by the interruption imbalance, a fundamental bottleneck inherent to the generation of universal perturbations. We reveal that conventional static gradient normalization fundamentally struggles to resolve architectural conflicts, causing the optimization to bias towards susceptible models while neglecting resistant ones. We argue that achieving high and uniform effectiveness requires resolving this imbalance by reaching an adaptive equilibrium. We propose the Adaptive Equilibrium Framework (AEF), which employs a dynamic weighting mechanism that utilizes real-time loss feedback to adaptively assign greater interruption weights to the most resistant models. This approach shifts the optimization from an average-case problem to finding a dynamic balance, driving the perturbation to a uniformly effective equilibrium state. Comprehensive experiments validate that AEF achieves a more balanced interruption performance, maintaining a consistent interruption success rate across the evaluated diverse architectures. Our code will be available at \url{https://anonymous.4open.science/r/AEF-1259/}.
\end{abstract}


% Keywords

\maketitle

% Main text
\section{Introduction}\label{}
In recent years, generative artificial intelligence, particularly deepfake technology, has made significant advancements \citep{b1}. These models generate highly realistic forgeries that attackers exploit for political disinformation, identity theft, and financial fraud. Such malicious applications compromise personal privacy and pose a serious threat to the social stability and credibility of public institutions \citep{b2,b11,b12}.

Strategies to deal with these threats include passive detection and active defense. Passive detection analyzes content to identify forgery artifacts \citep{b3,b4,b5,b44,b49,b50,b51}. It can be used to assess digital media without relying on the generative source or access to the original data. However, this method has limitations, as it cannot prevent forged content creation or real-time reputation damage. Therefore, active defense, which intervenes directly in the generation process, has emerged as a critical research direction \citep{b7,b8,b9,b10,b17,b25,b45,b46,b47,b48}. Common active defense paradigms include data poisoning \citep{b16,b22} and interruption defense\citep{b8,b9,b10,b17,b13,b14}. However, data poisoning is ineffective against pre-trained models. This limitation motivates the interruption defense, a paradigm designed to distort the deepfake output during the inference stage.
\begin{center}
    \centering
    \includegraphics[width=\linewidth]{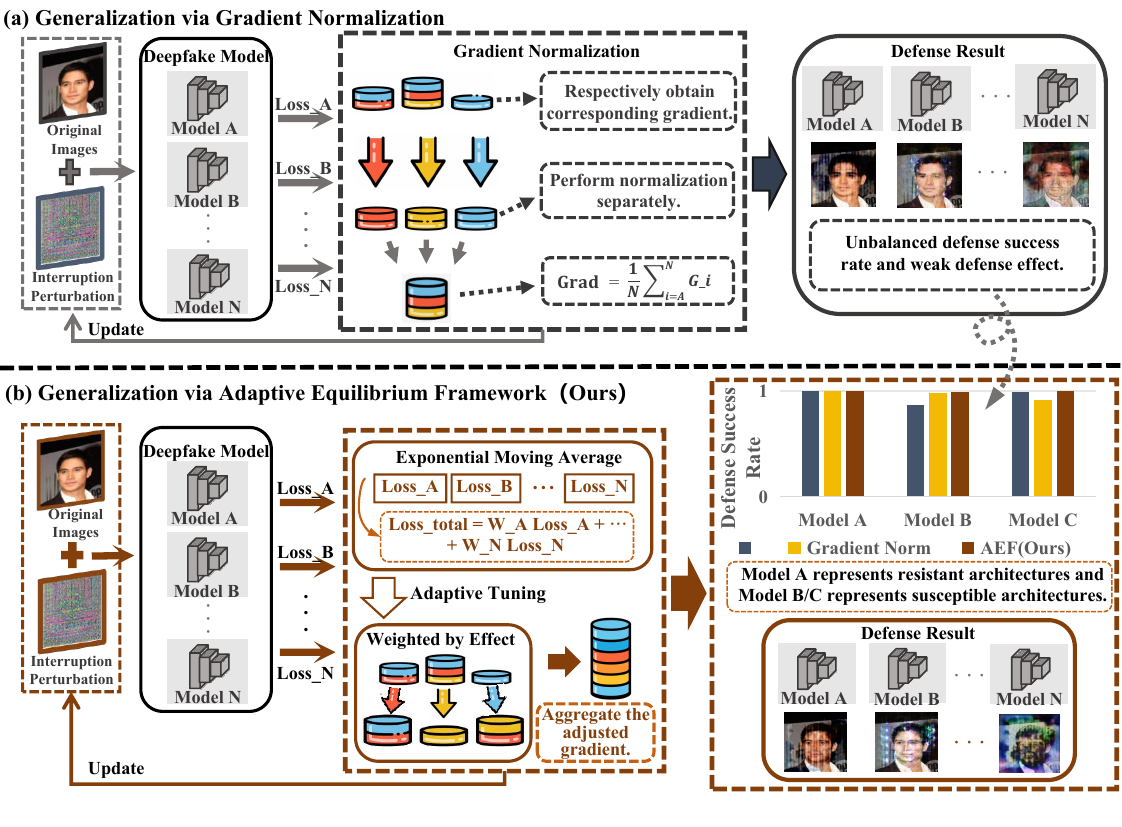}  
    \captionof{figure}{Illustration of two universal strategies for interruption defense. (a) illustrates the traditional generalization method based on gradient averaging. This strategy calculates the loss and gradient for each deepfake model separately, and then guides the integrated update of the perturbation through simple gradient averaging. (b) illustrates the proposed generalization method based on AEF. This strategy first calculates the loss for each model and utilizes the Exponential Moving Average to track historical interruption effectiveness. Subsequently, through Adaptive Tuning, it dynamically allocates weights based on the interruption effectiveness of each model. The resulting aggregated loss then guides the unified perturbation update.}
    \label{fig:01}
\end{center}
Initial interruption strategies are often limited to generating sample-specific perturbations that are only effective for a specific image against a specific model \citep{b10, b43}. Such dependency restricts their applicability, as optimizing a unique perturbation for every input image is infeasible for large-scale active defense. Consequently, the field has evolved into universal interruption perturbations, which exploit common vulnerabilities to generate a single perturbation. Representative works in this domain include CMUA-Watermark \citep{b15} and FOUND \citep{b20}.

Although universal interruption methods have achieved notable effects on multiple models, these methods typically rely on a static gradient normalization strategy as shown in Fig. \ref{fig:01} (a). This static approach ignores architectural differences between diverse models and causes conflicts when updating the ensemble gradient. And this strategy is often dominated by gradients from the susceptible models, limiting the cross-architecture transferability performance \citep{b40}.

To overcome this limitation, we propose a dynamic optimization framework based on adaptive loss weighting as shown in Fig. \ref{fig:01} (b). This framework dynamically adjusts each loss component's contribution and uses an dynamic weighting module to assign weights based on real-time interruption effectiveness. By utilizing an Exponential Moving Average (EMA) to track performance, the method shifts the optimization focus towards the most resistant models. This targeted optimization not only resolves gradient conflicts, but also achieves a more powerful and uniform cross-architecture transferability ability.

% Use the style of numbering in square brackets.
% If nothing is used, default style will be taken.
%\begin{enumerate}[a)]
%\item 
%\item 
%\item 
%\end{enumerate}  
The main contributions of this paper are summarized as follows:
\begin{itemize}
\item We propose AEF, which employs dynamic loss weighting to achieve adaptive equilibrium, addressing the imbalance problem inherent in universal interruption defense.\par
\item  We design a paradigm-aware feature disruption strategy that is introduced to boost feature-level interruption efficacy, improving the universal interruption effect. \par
\item Experimental results demonstrate that our method mitigates the performance imbalance across different deepfake models and produces more consistent and generalized interrupt efficiency than existing static strategies.
\end{itemize}

\section{Related Work}
\subsection{Architectural Diversity and Gradient Conflicts}
The architectural differences of four representative models are: StarGAN \citep{b36}, AttGAN \citep{b37}, AGGAN \citep{b35} and HiSD \citep{b34}.
Specifically, StarGAN concatenates the target attribute vector directly with the input image at the initial layer, causing the attribute signal to globally permeate the entire network and entangle with both the feature extraction and generation phases. In contrast, AttGAN adopts a decoupled strategy by injecting the attribute vector exclusively into the latent space; this ensures the encoder acts as an attribute-independent feature extractor, limiting the attribute's influence solely to the generator. To achieve more precise editing, AGGAN introduces an attention mechanism to generate spatial masks, deliberately confining the attribute manipulation to specific foreground regions and thereby creating a strictly localized optimization objective. Finally, moving beyond simple binary condition vectors, HiSD extracts a high-dimensional, complex style representation from a reference image and embeds it deep within the generator through a specialized feature mixing module.

When a static optimization paradigm attempts to find a universal perturbation $W$ by minimizing a simple average loss, it implicitly attempts to solve a deeply conflicted problem. The aggregated gradient is an average of four mathematically incompatible vectors, each targeting a unique model and mechanism. These vectors include a gradient from StarGAN optimized to disrupt an attribute signal at the input layer and a gradient from AttGAN targeting the feature compression layer. The aggregation also includes a gradient from AGGAN designed to disrupt a spatially-localized attention mask and a gradient from HiSD optimized to disrupt a style code at an intermediate generator layer. These gradients are structurally misaligned and represent different optimization directions. Consequently, optimization is inevitably dominated by the easiest model, leading to the imbalanced interruption problem \citep{b40}. This analysis establishes the clear technical necessity for a dynamic mechanism that can adaptively resolve these fundamental architectural conflicts.

\subsection{Comparison with Related Optimization Strategies}
The core idea of the AEF framework to dynamically focus optimization resources on more difficult instances is conceptually related to established strategies. Focal Loss \citep{b38} concentrates on difficult negative examples by reducing the weight of simple samples during training. OHEM \citep{b30} adaptively increases the sampling weights of high-loss data points.

However, the AEF is different from these strategies in its problem domain, optimization objective and execution mechanism.
First, there are differences in the areas and mechanisms. Both Focal Loss and OHEM are model training techniques designed to optimize the classifier parameters. They apply weighting to individual data samples within a training set. In sharp contrast, the AEF is an adversarial interruption strategy. Its goal is to find a universal perturbation vector. The weighting mechanism targets entire models within the ensemble, not individual samples. And these weights are determined by a model's historical EMA-smoothed loss, rather than by a single-instance classification result.
In addition, the optimization objectives are different. The goal of Focal Loss and OHEM is to improve the average classification accuracy of a model. However, the AEF introduces a new optimization objective termed Adversarial Equilibrium. Instead of merely maximizing the average interruption effect, this objective is designed to ensure that the perturbation achieves uniform destructive efficacy against all ensemble members, particularly the most resistant.

Therefore, rather than the invention of dynamic weighting itself, the core contribution of AEF resides in its pioneering application to the universal adversarial interruption domain, establishing a dedicated framework that systematically resolves the critical interruption imbalance bottleneck fundamentally rooted in model architectural conflicts.

% Uncomment and use as the case may be
%\begin{theorem} 
%\end{theorem}

% Uncomment and use as the case may be
%\begin{lemma} 
%\end{lemma}

\vspace{1em} % 上方留一点空白，美观一点
\begin{figure*}[t]
    \centering
    \includegraphics[scale=0.74 ]{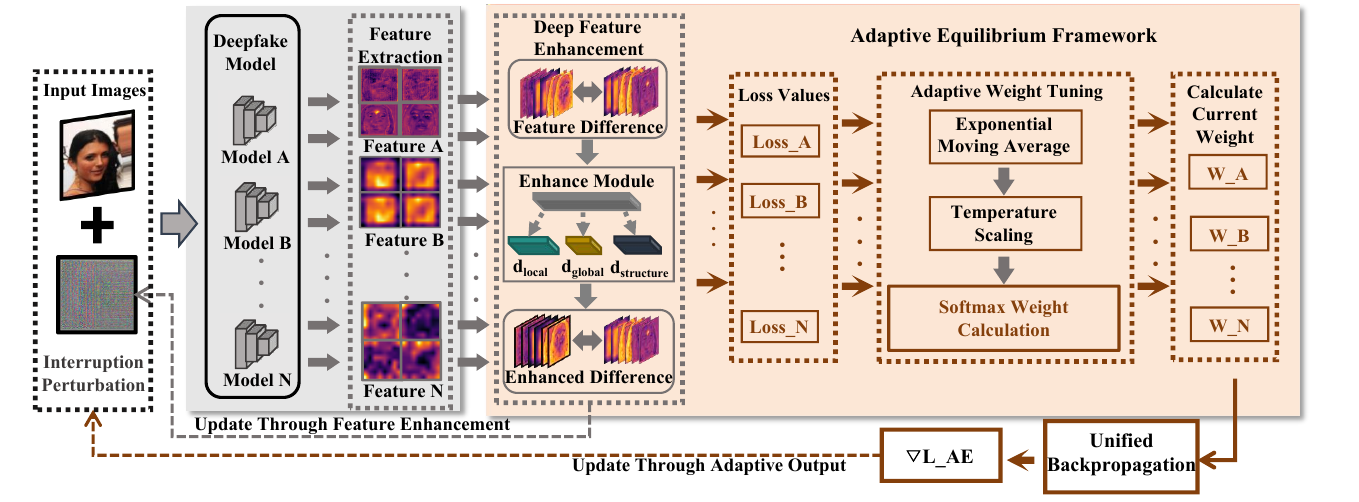}  
    \caption{Illustration of the proposed Adaptive Equilibrium Framework.}
    \label{fig:02}
\end{figure*}
\vspace{1em} % 下方留一点空白
\vspace{-3em}
\section{Method}\label{}

This section details the AEF, with its architecture illustrated in Fig. \ref{fig:02}. The framework optimizes a perturbation through two concurrent branches. The Deep Feature Enhancement  (DFE) branch aims to corrupt semantic information by enhancing the variance between extracted feature representations, generating a feature-level update. Concurrently, the Adaptive Weight Tuning dynamically calculates each model's contribution. It stabilizes each model loss using the EMA and normalizes them into a probability distribution. Critically, these weights are applied to their corresponding losses to construct a unified total loss. This total loss is then backpropagated once to produce the adaptive update gradient ($\nabla L_\_{AE}$), which is combined with the feature update to refine the perturbation.
\subsection{Deep Feature Enhancement}
Instead of relying on a single overall distance metric, generating universal perturbations across various deepfake architectures requires a multi-dimensional strategy. To construct an effective feature-level disruption, the DFE module decouples the difference between clean ($F_{clean}$) and adversarial ($F_{adv}$) features into three components, each targeting a specific generative mechanism:

\textbf{Local Pattern Discrepancy ($\mathbf{d}_{local}$):} Style-guided generators (e.g., HiSD) heavily rely on instance-level statistics for adaptive style injection. We compute this discrepancy using Instance Normalization $\text{IN}(\cdot)$ to explicitly corrupt the localized texture representations essential for structural styling:
\vspace{-0.3em}
\begin{equation}
\mathbf{d}_{local} = \text{IN}(F_{adv}) - \text{IN}(F_{clean}).
\end{equation}

\textbf{Global Statistical Discrepancy ($\mathbf{d}_{global}$):} Attribute-conditioned networks (e.g., AttGAN, StarGAN) manipulate global latent representations. To disrupt this mechanism, we capture the distribution shift by calculating the normalized difference in global mean ($\mu$) and standard deviation ($\sigma$), forcing a severe departure from the original semantic distribution:
\begin{equation}
\mathbf{d}_{global} = \frac{\mu(F_{adv}) - \mu(F_{clean})}{\sigma(F_{clean}) + \epsilon}.
\end{equation}

\textbf{Structural Semantic Discrepancy ($\mathbf{d}_{structure}$):} To blind attention-guided architectures (e.g., AGGAN) that depend on localized correlation maps, we utilize Channel Self-Attention $\text{CSA}(\cdot)$. This component severs the critical channel-wise dependencies, causing severe feature fragmentation:
\begin{equation}
\mathbf{d}_{structure} = \text{CSA}(F_{adv}) - \text{CSA}(F_{clean}).
\end{equation}

Unlike approaches TSDF \citep{b42} that use non-linear exponential functions to amplify feature discrepancies, AEF avoids such complexities. Exponentially-driven losses are structurally incompatible with our Adaptive Weight Tuning module because they induce fluctuations in the EMA, which prevents the system from maintaining a stable adaptive equilibrium. Therefore, we design a maximization strategy. The final feature loss $L_{\text{feat}}$ aims to maximize these decoupled discrepancies. It is defined as the weighted sum of their L2 norms, where $k \in \{\text{local, global, structure}\}$:
\begin{equation}
L_{\text{feat}} = - \sum_{k} w_k \cdot \|\mathbf{d}_k\|_2 .
\label{eq:4}
\end{equation}

\subsection{Adaptive Equilibrium Mechanism}

After establishing a stronger perturbation signal based on the DFE module, we further addressed the issues of generalization and balance using Adaptive Equilibrium Framework. This mechanism, as detailed in Fig. \ref{fig:02}, allocates optimization resources to ensure uniform interruption efficacy across the entire set of models.

Multi-Level Interrupt Objective:
For each model $i$ in the ensemble, we define a composite loss $L_{total}^{(i)}$ that provides a comprehensive evaluation of the interruption performance. This function is a weighted sum of an end-to-end output loss $L_{e2e}$ and the intermediate feature-level loss $L_{feat}$.

\begin{equation}
L_{total}^{(i)} = (1-\lambda) L_{e2e}^{(i)} + \lambda L_{feat}^{(i)} , \label{eq:5}
\end{equation}
where $L_{e2e}^{(i)}$ and $L_{feat}^{(i)}$ are defined to maximize the L2 distance between adversarial and clean samples at the output and feature levels, respectively. The hyperparameter $\lambda$ balances their contributions. It ensures that optimization considers interruption at both the model's internal and final output stages.

Performance Smoothing with EMA:
Because loss values from a single iteration can be volatile, we employ Exponential Moving Average (EMA) to maintain a smoothed historical record of the composite loss for each model:
\begin{equation}
L_{ema}^{(t, i)} = \beta L_{ema}^{(t-1, i)} + (1-\beta) L_{total}^{(t, i)},\label{eq:6}
\end{equation}
where $L_{ema}^{(t, i)}$ is the smooth loss for model $i$ in iteration $t$, and $\beta$ is the smoothing factor. A higher value of $L_{ema}^{(i)}$ means that the interruption is consistently less effective against the model $i$.

Dynamic Weight Allocation with Temperature-Controlled Softmax:
The vector of smoothed losses $\mathbf{L}_{ema}$ is then fed into a temperature-controlled Softmax function to compute dynamic weights, $w_i$, for each model in the next optimization step:
\begin{equation}
w_i = \frac{\exp(L_{ema}^{(i)} / T)}{\sum_{j} \exp(L_{ema}^{(j)} / T)}  ,
\label{eq:7}
\end{equation}
where $T$ is a temperature hyperparameter. This mechanism is the core of our adaptive solution. The Softmax function assigns higher weights to models with higher smoothed losses. The temperature $T$ controls the sharpness of this allocation. As $T \to 0$, the mechanism focuses almost exclusively on the single most resistant model, while a larger $T$ results in a more distributed allocation.

Unified Weighted Backpropagation:
Finally, the composite losses of all models are aggregated into a single global loss $L_{global}$, through a weighted sum using the dynamically computed weights $w_i$:
\begin{equation}
L_{global} = \sum_{i} w_i L_{total}^{(i)} .\label{eq:8}
\end{equation}

A single backpropagation is performed on this $L_{global}$. The resulting gradient is a weighted average, predominantly guided by the models that are currently the most challenging. This process forces the optimization to find a perturbation that achieves an equilibrium of effectiveness, overcoming the issue of imbalanced interruption performance. The overall pipeline of the introduced method is detailed in Algorithm \ref{alg:aef_final}.

\begin{algorithm}[h!]
\caption{Adaptive Equilibrium Framework (AEF)}
\label{alg:aef_final}
\begin{algorithmic}[1]
    \STATE \textbf{Input:} Image batches $X$, model ensemble $\{M_i\}$.
    \STATE \textbf{Parameter:} Outer iterations $T_{out}$, inner iterations $T_{in}$.
    \STATE \textbf{Output:} Universal perturbation $W$.
    \STATE \COMMENT{Update($\delta, \mathcal{L}$) denotes a gradient-based update step, e.g., MI-FGSM.}
    \STATE Initialize $W$ randomly.
    \STATE Initialize EMA loss vector $\mathbf{L}_{ema} \leftarrow \mathbf{0}$.
    \STATE \textbf{for} $t = 1$ \textbf{to} $T_{out}$ \textbf{do}
    \STATE \quad \textbf{for} each batch $x_b$ \textbf{in} $X$ \textbf{do}
    \STATE \quad \quad \textit{// Stage 1: Feature Enhancement }
    \STATE \quad \quad \textbf{for} $j = 1$ \textbf{to} $T_{in}$ \textbf{do}
    \STATE \quad \quad \quad $L_{\text{feat}} \leftarrow$ Compute aggregated loss (Eq. \ref{eq:4}.
    \STATE \quad \quad \quad $\delta \leftarrow \text{Update}(\delta, L_{\text{feat}})$.
    \STATE \quad \quad \textbf{end for}
    \STATE \quad \quad \textit{// Stage 2: Adaptive Equilibrium}
    \STATE \quad \quad $L_{total} \leftarrow$ Compute comprehensive loss vector for all models (Eq. \ref{eq:5}).
    \STATE \quad \quad $L_{ema} \leftarrow$ Update with EMA via Eq. \ref{eq:6}.
    \STATE \quad \quad $\mathbf{w} \leftarrow$ Compute adaptive weights via Eq. \ref{eq:7}.
    \STATE \quad \quad $L_{global} \leftarrow \mathbf{w} \cdot \mathbf{L}_{total}$ \COMMENT{Weighted global loss (Eq. \ref{eq:8})}
    \STATE \quad \quad $\delta \leftarrow \text{Update}(\delta, \mathcal{L}_{global})$.
    \STATE \quad \textbf{end for}
    \STATE \textbf{end for}
    \STATE \textbf{return} $W$.
\end{algorithmic}
\end{algorithm}

\begin{table}[h]
    \centering
    \caption{Comparison of Interruption Performance among CMUA, FOUND, DWT, TSDF and AEF. The avg is the average result of each indicator under the dataset. The best result is marked in bold.}
    \label{tab:table2} % 使用了新的label
    \setlength{\tabcolsep}{2.3pt}
    \renewcommand{\arraystretch}{1.2} % 调整行高
    \scriptsize % 如果需要，可以取消注释
    \begin{tabular}{ l l l c c c c c }
        
        \hline
        Dataset & Model & Method & L2mask$\uparrow$ & SRmask$\uparrow$ & FID$\uparrow$ & PSNR$\downarrow$ & SSIM$\downarrow$ \\
        \hline

        % --- CelebA Block ---
        \multirow{25}{*}{CelebA} & \multirow{5}{*}{StarGAN} 
            & CMUA  & 0.21 & \ 100.00\% & 305.05 & 12.85 & 0.48 \\
            & & FOUND & 0.26 & 100.00\% & 362.54 & 11.45 & 0.26 \\
            & & DWT   & 0.18 & 100.00\% & 220.45 & 15.31 & 0.53 \\
            & & TSDF  & 0.37 & 100.00\% & 378.89 & 9.73 & 0.17 \\
            & & AEF   & \textbf{0.38} & \textbf{100.00\%} & \textbf{380.71} & \textbf{9.66} & \textbf{0.13} \\
        \cline{2-8}
        & \multirow{5}{*}{AGGAN} 
            & CMUA  & 0.23 & 99.93\% & 163.00 & 15.93 & 0.62 \\
            & & FOUND & 0.19 & 100.00\% & 224.97 & 16.12 & 0.68 \\
            & & DWT   & 0.15 & 100.00\% & 140.38 & 21.40 & 0.77 \\
            & & TSDF  & 0.18 & 100.00\% & 221.36 & 17.86 & 0.70 \\
            & & AEF   & \textbf{0.18} & \textbf{100.00\%} & \textbf{230.75} & \textbf{17.40} & \textbf{0.68} \\
        \cline{2-8}
        & \multirow{5}{*}{AttGAN} 
            & CMUA  & 0.13 & 87.01\% & 180.95 & 18.59 & 0.69 \\
            & & FOUND & 0.15 & 97.90\% & 183.22 & 17.36 & 0.64 \\
            & & DWT   & 0.01 & 2.10\% & 54.19 & 31.69 & 0.95 \\
            & & TSDF  & 0.17 & 98.25\% & 192.31 & 16.76 & 0.62 \\
            & & AEF   & \textbf{0.21} & \textbf{99.65\%} & \textbf{218.23} & \textbf{15.35} & \textbf{0.55} \\
        \cline{2-8}
        & \multirow{5}{*}{HiSD} 
            & CMUA  & \textbf{0.21} & 99.72\% & \textbf{188.85} & 16.09 & 0.75 \\
            & & FOUND & 0.09 & 92.55\% & 183.46 & 19.11 & 0.82 \\
            & & DWT   & 0.03 & 7.62\% & 148.29 & 26.45 & 0.87 \\
            & & TSDF  & 0.17 & 99.15\% & 170.99 & 16.14 & 0.76 \\
            & & AEF   & 0.17 & \textbf{99.85\%} & 186.81 & \textbf{15.56} & \textbf{0.74} \\
        \cline{2-8}
        & \multirow{5}{*}{Avg} 
            & CMUA  & 0.20 & 96.67\% & 209.46 & 15.87 & 0.64 \\
            & & FOUND & 0.17 & 97.61\% & 238.55 & 16.01 & 0.60 \\
            & & DWT   & 0.09 & 52,45\% & 140.82 & 23.71 & 0.78 \\
            & & TSDF  & 0.22 & 99.35\% & 240.88 & 15.12 & 0.56 \\
            & & AEF   & \textbf{0.24} & \textbf{99.88\%} & \textbf{254.12} & \textbf{14.49} & \textbf{0.52} \\
        \hline
    \end{tabular}
\end{table}
\begin{table}[H]
    \centering
    \addtocounter{table}{-1} % 关键修改1：将表格编号计数器减1，使其与上表保持同号
    \caption{Comparison of Interruption Performance among CMUA, FOUND, DWT, TSDF and AEF (continued)} % 关键修改2：添加(continued)标识
    \setlength{\tabcolsep}{2.3pt} 
    \renewcommand{\arraystretch}{1.2} % 调整行高
    \scriptsize % 如果需要，可以取消注释
    \begin{tabular}{ l l l c c c c c }
        \hline
        % 关键修改3：为续表重新加上表头，否则换页后读者无法对照列名
        Dataset & Model & Method & L2mask$\uparrow$ & SRmask$\uparrow$ & FID$\uparrow$ & PSNR$\downarrow$ & SSIM$\downarrow$ \\
        \hline
        % --- LFW Block ---
        \multirow{25}{*}{LFW} & \multirow{5}{*}{StarGAN} 
            & CMUA  & 0.21 & \ 100.00\% & 298.86 & 12.92 & 0.47 \\
            & & FOUND & 0.26 & 100.00\% & \textbf{398.86} & 15.86 & 0.27 \\
            & & DWT   & 0.17 & 100.00\% & 287.46 & 15.39 & 0.56 \\
            & & TSDF  & 0.36 & 100.00\% & 372.80 & 10.32 & 0.20 \\
            & & AEF   & \textbf{0.38} & \textbf{100.00\%} & 366.45 & \textbf{10.18} & \textbf{0.15} \\
        \cline{2-8}
        & \multirow{5}{*}{AGGAN} 
            & CMUA  & \textbf{0.24} & \ 100.00\% & 193.57 & 16.10 & \textbf{0.61} \\
            & & FOUND & 0.20 & 100.00\% & 239.06 & \textbf{15.72} & 0.66 \\
            & & DWT   & 0.15 & 100.00\% & 169.30 & 21.20 & 0.78 \\
            & & TSDF  & 0.19 & 100.00\% & 228.19 & 17.97 & 0.69 \\
            & & AEF   & 0.20 & \textbf{100.00\%} & \textbf{241.97} & 17.49 & 0.66 \\
        \cline{2-8}
        & \multirow{5}{*}{AttGAN} 
            & CMUA  & 0.11 & 95.24\% & 207.76 & 18.51 & 0.67 \\
            & & FOUND & 0.16 & 98.05\% & 248.99 & 16.70 & 0.60 \\
            & & DWT   & 0.01 & 2.16\% & 77.43 & 30.72 & 0.94 \\
            & & TSDF  & 0.16 & 99.01\% & 253.76 & 16.17 & 0.59 \\
            & & AEF   & \textbf{0.22} & \textbf{99.84\%} & \textbf{292.45} & \textbf{14.62} & \textbf{0.51} \\
        \cline{2-8}
        & \multirow{5}{*}{HiSD} 
            & CMUA  & 0.12 & 97.44\% & \textbf{221.82} & 16.22 & 0.75 \\
            & & FOUND & 0.07 & 91.04\% & 163.23 & 19.61 & 0.82 \\
            & & DWT   & 0.04 & 16.07\% & 200.82 & 26.01 & 0.84 \\
            & & TSDF  & 0.12 & 95.42\% & 188.98 & 16.58 & 0.76 \\
            & & AEF   & \textbf{0.14} & \textbf{98.64\%} & 175.27 & \textbf{15.90} & \textbf{0.74} \\
        \cline{2-8}
        & \multirow{5}{*}{Avg} 
            & CMUA  & 0.17 & 98.17\% & 230.50 & 15.93 & 0.63 \\
            & & FOUND & 0.17 & 97.27\% & 247.53 & 16.97 & 0.59 \\
            & & DWT   & 0.09 & 54.56\% & 158.75 & 23.33 & 0.78 \\
            & & TSDF  & 0.21 & 98.61\% & 260.93 & 15.26 & 0.56 \\
            & & AEF   & \textbf{0.24} & \textbf{99.62\%} & \textbf{261.53} & \textbf{14.72} & \textbf{0.52} \\
        \hline
        \multirow{25}{*}{FF++O} & \multirow{5}{*}{StarGAN} 
            & CMUA  & 0.21 & 100.00\% & 336.31 & 13.58 & 0.55 \\
            & & FOUND & 0.24 & 100.00\% & 380.27 & \textbf{8.49} & 0.37 \\
            & & DWT   & 0.17 & 100.00\% & 243.90 & 16.00 & 0.62 \\
            & & TSDF  & 0.35 & 100.00\% & 400.15 & 10.74 & 0.23 \\
            & & AEF   & \textbf{0.38} & \textbf{100.00\%} & \textbf{409.29} & 10.10 & \textbf{0.14} \\
        \cline{2-8}
        & \multirow{5}{*}{AGGAN} 
            & CMUA  & \textbf{0.24} & 100.00\% & 276.09 & 15.14 & 0.70 \\
            & & FOUND & 0.22 & 100.00\% & \textbf{370.10} & \textbf{13.25} & 0.77 \\
            & & DWT   & 0.15 & 100.00\% & 156.50 & 21.08 & 0.78 \\
            & & TSDF  & 0.19 & 100.00\% & 299.18 & 17.66 & 0.74 \\
            & & AEF   & 0.19 & \textbf{100.00\%} & 265.94 & 17.15 & \textbf{0.70} \\
        \cline{2-8}
        & \multirow{5}{*}{AttGAN} 
            & CMUA  & 0.14 & 91.80\% & 216.97 & 15.71 & 0.70 \\
            & & FOUND & 0.21 & 99.51\% & \textbf{282.66} & 16.81 & 0.60 \\
            & & DWT   & 0.01 & 2.90\% & 67.37 & 30.87 & 0.95 \\
            & & TSDF  & 0.16 & 99.02\% & 260.94 & 15.47 & 0.57 \\
            & & AEF   & \textbf{0.21} & \textbf{99.85\%} & 264.91 & \textbf{14.96} & \textbf{0.53} \\
        \cline{2-8}
        & \multirow{5}{*}{HiSD}
            & CMUA  & 0.11 & 86.15\% & \textbf{213.57} & 17.99 & 0.83 \\
            & & FOUND & 0.09 & 85.35\% & 178.71 & 17.10 & 0.88 \\
            & & DWT   & 0.03 & 11.32\% & 154.38 & 26.72 & 0.89 \\
            & & TSDF  & 0.11 & 94.95\% & 176.98 & 16.67 & 0.81 \\
            & & AEF   & \textbf{0.14} & \textbf{98.62\%} & 173.95 & \textbf{15.48} & \textbf{0.77} \\
        \cline{2-8}
        & \multirow{5}{*}{Avg} 
            & CMUA  & 0.18 & 97.23\% & 270.99 & 15.61 & 0.70 \\
            & & FOUND & 0.19 & 97.71\% & \textbf{305.44} & \textbf{13.91} & 0.66 \\
            & & DWT   & 0.09 & 53.56\% & 155.54 & 23.67 & 0.82 \\
            & & TSDF  & 0.20 & 98.49\% & 284.31 & 15.14 & 0.59 \\
            & & AEF   & \textbf{0.23} & \textbf{99.61\%} & 278.52 & 14.42 & \textbf{0.54} \\
        \hline 
        
    \end{tabular}
\end{table}
\vspace{-1em}
\section{Experiments}
\subsection{Experimental Settings}
\subsubsection{Datasets}
For the generation of our universal perturbation, we utilize the aligned and cropped version of the CelebA dataset \citep{b31}. Following the efficient training paradigm established by the SOTA works \citep{b20}, we also employ a compact subset of 128 images to optimize the perturbation. This concise set is sufficient to capture the generalizable vulnerabilities of facial structures. The optimized perturbation is applied to the entire CelebA test split. To rigorously assess the cross-dataset generalization of AEF, we perform additional tests on two unseen datasets: Labeled Faces in the Wild (LFW) \citep{b32} and facial images extracted from the original videos of the FaceForensics++ (FF++O) dataset \citep{b33}.

\subsubsection{Deepfake Models}
To evaluate our adaptive framework, the experiments utilize four representative facial attribute editing models: StarGAN\citep{b36}, AttGAN\citep{b37}, AGGAN\citep{b35} and HiSD\citep{b34}, all of which are trained on the CelebA dataset.

\subsubsection{Baseline}
To validate the superiority of our framework, we benchmark it against a comprehensive set of SOTA methods. These include foundational universal interruption strategies like CMUA-Watermark \citep{b15} and FOUND \citep{b20}, the DWT-based defense framework \citep{b41} which employs an alternative dynamic weighting scheme, and the persistence-focused TSDF \citep{b42} which combines interruption with poisoning. This comparative analysis quantitatively demonstrates our framework's superior performance across an extensive array of metrics. This success is a direct result of achieving its core optimization objectives.\\

\subsubsection{Implementation Details}

Our hyperparameter configuration is established to jointly optimize the framework's synergistic components. The universal perturbation is limited within an $L_\infty$ norm of 0.05 to maintain visual fidelity while ensuring interruption efficacy. The framework is optimized for 30 total iterations, which we found to be the optimal point for performance to converge without incurring unnecessary computational overhead. We adopt the feature shift hyperparameter $\alpha$ set at 0.8. Key hyperparameters for the adaptive mechanism are determined through ablation studies. The loss balance factor $\lambda$ is set to 0.001. The EMA smoothing factor $\beta$ is 0.9 and the temperature $T$ for the Softmax weighting is set to 0.1.

\subsubsection{Evaluation Metrics}
Defense Effectiveness Metrics: We quantify the interruption efficacy using conventional standard metrics. Following previous work \citep{b20}, we define a successful disruption where the $L2mask > 0.05$. Therefore, SRmask represents the percentage of images that meet the standard, and higher values for both indicators signify a more effective interruption. We also employ the Fréchet Inception Distance (FID) to measure the perceptual dissimilarity between the interrupted outputs and the original images. A higher FID signifies a stronger interrupted effect. Additionally, we use the Peak Signal-to-Noise Ratio (PSNR) and the Structural Similarity Index (SSIM). Both metrics evaluate the fidelity discrepancy between the interrupted output and the original image, where lower scores indicate a more successful interruption. 

Image Quality Metrics: We evaluate the imperceptibility of our generated perturbation by quantifying the perceptual quality of protected images against the original images. This assessment employs the PSNR and the SSIM. Higher values for both metrics signify minimal perceptual loss, thus confirming the perturbation's imperceptibility. Additionally, the perturbation's utility in downstream applications is assessed, specifically face anti-spoofing (HyperFAS) and face recognition (FaceNet \citep{b28}). This assessment compares performance on clean images with those modified by AEF and baselines.

\subsection{Compare the SOTA Methods}

\subsubsection{Interruption Performance Against SOTA Methods}
We evaluate the interrupted performance of AEF against CMUA, FOUND, DWT and TSDF with the detailed quantitative results presented in Table \ref{tab:table2}. The results demonstrate the clear superiority of AEF, which causes more significant perceptual destruction to the generated images.

This superior performance is particularly evident in our method's ability to corrupt critical facial regions, as measured by the SRmask metric. Our framework achieves a nearly perfect average of 99.88\% and maintains scores exceeding 99.50\% on challenging architectures such as AttGAN and HiSD. This high efficiency is further supported by our framework's performance across other evaluation metrics, where it also obtains higher average FID values and lower average SSIM scores. These results signify a greater destruction of both perceptual quality and structural integrity compared to the baselines.
Furthermore, the framework's superior performance is not confined to the average metrics. This robust performance is consistently maintained across all tested datasets, including CelebA, LFW, and FF++. For instance, on the LFW and FF++ datasets, AEF achieves SRmask of 99.62\% and 99.61\%, consistently surpassing the baseline methods. The AEF also consistently exhibits a more powerful disruptive effect on the individual models tested, which confirms the comprehensive nature of its capacity to degrade output quality and integrity.
\vspace{-3em}
\vspace{1em} % 上方留一点空白，美观一点
\begin{figure*}[t]
    \centering
    \includegraphics[scale=0.3]{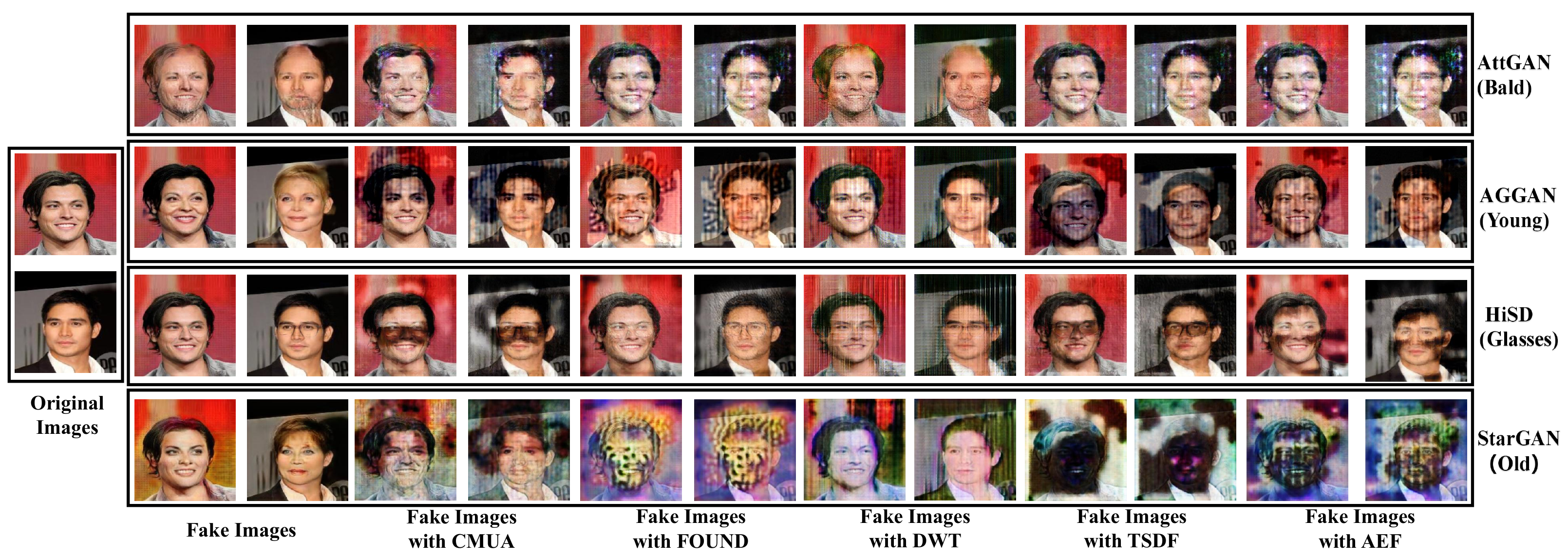}  
    \caption{This figure extends the qualitative analysis to the CelebA dataset.}
    \label{fig:03}
\end{figure*}
\vspace{1em} % 下方留一点空白
\vspace{1em} % 上方留一点空白，美观一点
\begin{figure*}[t] 
    \centering
    \includegraphics[scale=0.3]{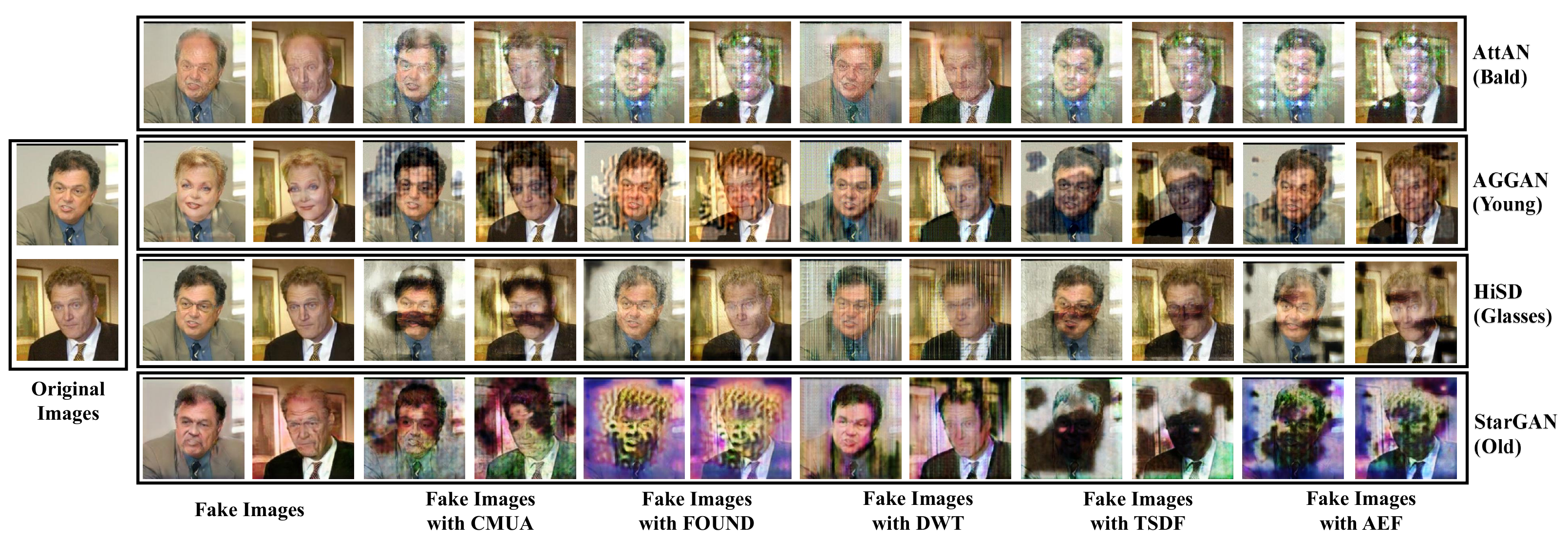}  
    \caption{This figure extends the qualitative analysis to the LFW dataset.}
    \label{fig:04}
\end{figure*}
\vspace{1em} % 下方留一点空白
\vspace{1em} % 上方留一点空白，美观一点
\begin{figure*}[t]
    \centering
    \includegraphics[scale=0.3]{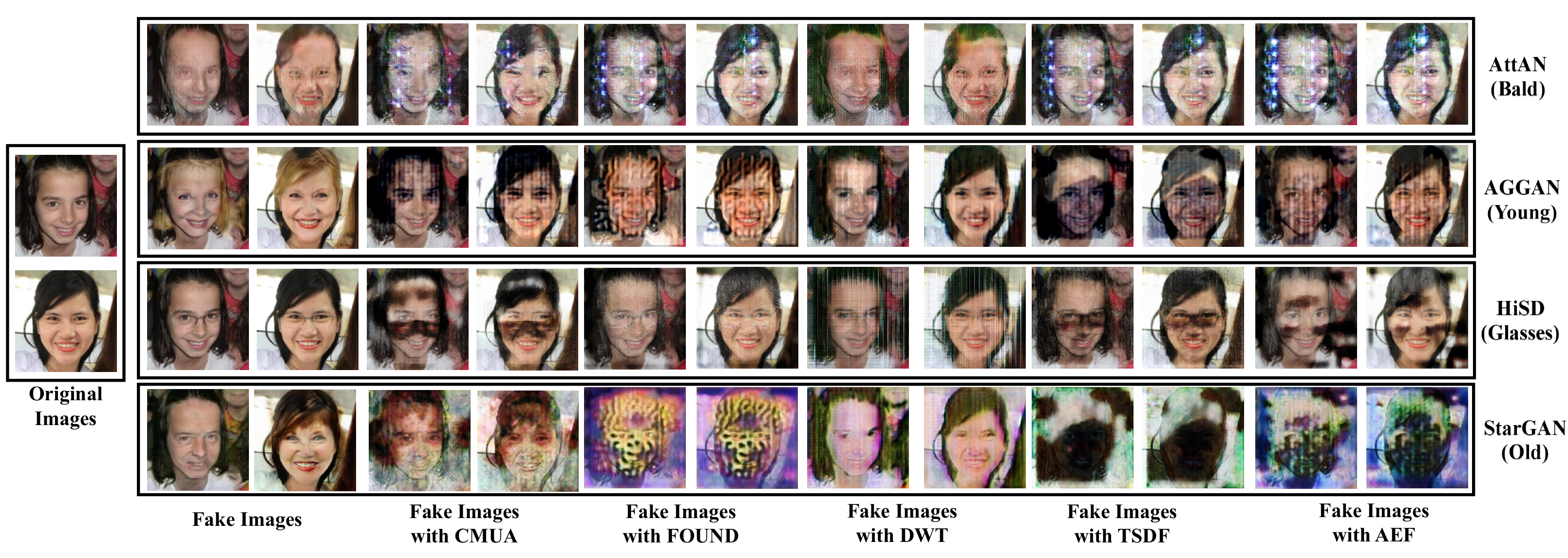}  
    \caption{This figure extends the qualitative analysis to the FF++O dataset.}
    \label{fig:05}
\end{figure*}
\vspace{1em} % 下方留一点空白
\vspace{-3em}
\subsubsection{Qualitative Analysis of Interruption Performance}

As shown in Fig. \ref{fig:03}, a qualitative evaluations show that AEF causes more significant structural degradation in the generated output compared to baselines. It achieves more substantial and structural destruction, causing forgery failures such as incoherent noise or feature fragmentation on models like StarGAN and AGGAN. In contrast, baseline defenses primarily introduce superficial color distortions or grid-like artifacts, leaving greater recognizable. This advantage persists on models like HiSD and AGGAN, where AEF generates widespread structural artifacts instead of the localized distortions seen from the baselines. Visual evidence confirms AEF achieves stronger interruption by compromising deep generative representations, not just superficial image quality.

To further demonstrate that this structural destruction is not limited to a specific data distribution, qualitative evaluations are extended to the LFW and FF++O datasets. As shown in Fig. \ref{fig:04} and Fig. \ref{fig:05}, AEF’s structural destruction maintains a consistent disruptive efficacy across diverse facial distributions. Specifically, on LFW, AEF achieves substantial structural destruction. In contrast, baseline defenses primarily introduce superficial vertical grid-like artifacts or localized color distortions, leaving the facial geometry highly recognizable. This objective advantage persists on the FF++O, where AEF continues to generate widespread structural artifacts instead of the localized dark smudges seen from the baselines. These visual evidences confirm that AEF achieves stronger interruption by fundamentally compromising deep generative representations across varying data distributions.

\subsubsection{Computational Efficiency Analysis}
To evaluate computational efficiency, we measured the time required to generate the universal perturbation on a single NVIDIA GeForce RTX 4090 GPU. As detailed in Table \ref{tab:table9}, AEF completes the optimization process in approximately 0.23 hours. This represents a reduction in training duration compared to the evaluated baselines under identical hardware conditions, requiring less time than FOUND (0.67 hours) and CMUA (over 5 hours). These results confirm that the proposed framework decreases the overall computational overhead for generating universal perturbations while maintaining the interruption efficacy.

\subsubsection{Perturbation Imperceptibility}
A primary design objective is ensuring the perturbation remains visually imperceptible. The quantitative results in Table \ref{tab:table5} confirm this low perceptibility. AEF consistently achieves higher SSIM and PSNR than baseline methods across all tested datasets, obtaining an SSIM of 0.91 on CelebA. These high scores confirm that our framework provides effective defense without degrading the perceptual quality of protected images.

The perturbation's impact on downstream utility is assessed using two representative tasks. As detailed in Table \ref{tab:downstream_performance}, the performance of these models on AEF-perturbed images is compared with the results from clean images and baseline-perturbed images. The Average Confidence Score (ACS) confirms that the AEF perturbation has a minimal impact on these applications, showing only a slight performance deviation from the original image.

\subsubsection{Evaluations in Black-Box Settings}
The primary objective of universal interruption frameworks, including FOUND, CMUA, and our proposed AEF, is to generate a single adversarial perturbation capable of disrupting multiple ensemble white-box models while generalizing to unseen images (data-level black-box). Since these frameworks are optimized on a highly restricted training subset, all evaluation images inherently represent a black-box setting. It is worth noting that the fundamental challenge addressed in this work is resolving optimization conflicts among multiple white-box source models; cross-architecture black-box transferability is not the primary design target.

To rigorously assess cross-model transferability and optimization stability, we conduct hold-out black-box evaluations (Table \ref{tab:hold_out_experiments}) by configuring three architectures (*) as the white-box ensemble and the remaining one as the unseen black-box target. Quantitative results reveal that the baseline CMUA suffers from internal optimization conflicts, causing its white-box performance to degrade, whereas FOUND exhibits limited black-box transferability despite maintaining white-box stability. In contrast, the proposed AEF demonstrates superior generalization equilibrium. By effectively mitigating gradient conflicts, AEF not only preserves a nearly 100\% defense success rate across all white-box source models but also achieves the highest SRmask and L2mask scores on almost all black-box targets (e.g., HiSD and AttGAN), objectively anchoring a more universally destructive adversarial representation in the feature space.

\subsubsection{Evaluation of Black-Box Transferability Driven by a Single Source Model}
To further investigate the cross-model generalization capabilities of defense algorithms under strictly limited prior knowledge, Table \ref{tab:single_source_black_box} evaluates the black-box transferability of adversarial perturbations optimized solely on StarGAN and tested against other unseen models. For a fair comparison, baseline methods are similarly restricted to single-model optimization on StarGAN. Quantitative results indicate that due to the high architectural homology in the underlying generation pipelines between AGGAN and the StarGAN, all evaluated methods maintain substantial disruption efficacy on AGGAN. However, when deployed against cross-architecture models with significantly divergent generation mechanisms, such as HiSD, traditional baselines exhibit clear limitations in black-box transferability. In contrast, our proposed method demonstrates better cross-architecture effect. This demonstrates that by maximizing the structural and semantic feature differences in the latent space, our design effectively mitigates the risk of the perturbation overfitting to specific network weights, thereby giving the defense signal a more generalized transferability from the fundamental feature dimension.
\begin{table}[htbp] 
\centering
\caption{Training Time Efficiency.}
\setlength{\tabcolsep}{5pt}
\renewcommand{\arraystretch}{1.2}
\begin{tabular}{c c c c c c}
\hline 
Method & CMUA & FOUND & DWT & TSDF & AEF\\ 
\hline
Time & $>$5h & $\approx$ 0.67h & $\approx$ 1.21h & $\approx$ 0.89h & $\approx$ \textbf{0.23h}\\
\hline 
\end{tabular}
\label{tab:table9}
\end{table}
\begin{table}[htbp]
    \centering
    \caption{Quantitative Comparison of Imperceptibility on Various Datasets.}
    \setlength{\tabcolsep}{4pt}
    \renewcommand{\arraystretch}{1.2}
    % 更改了 tabular 定义，添加了 |
    \begin{tabular}{l c c c c c c}
        \hline
        \multirow{2}{*}{Dataset} & \multirow{2}{*}{Metric} 
        % 为 \multicolumn 添加了 |
        & \multicolumn{5}{c}{Methods} \\
        \cline{3-7}
        & & CMUA & FOUND & DWT & TSDF & AEF \\
        \hline
        \multirow{2}{*}{CelebA} 
        & SSIM$\uparrow$ & 0.88 & 0.89 & 0.88 & 0.89 & \textbf{0.91} \\
        & PSNR$\uparrow$ & 32.45 & 33.03 & 32.72 & 33.07 & \textbf{33.17} \\
        \hline
        \multirow{2}{*}{LFW} 
        & SSIM$\uparrow$ & 0.88 & 0.89 & 0.88 & 0.89 & \textbf{0.90} \\
        & PSNR$\uparrow$ & 32.49 & 33.07 & 32.70 & 33.09 & \textbf{33.19} \\
        \hline
        \multirow{2}{*}{FF++O} 
        & SSIM$\uparrow$ & 0.87 & 0.88 & 0.88 & 0.89 & \textbf{0.90} \\
        & PSNR$\uparrow$ & 32.48 & 33.06 & 32.66 & 33.08 & \textbf{33.10} \\
        \hline
    \end{tabular}
    \label{tab:table5}
\end{table}
\begin{table}[h]
\centering
\caption{Impact of universal Perturbations on Downstream Model Performance.}
\label{tab:downstream_performance}
\setlength{\tabcolsep}{5pt} % Adjust column spacing if needed
\renewcommand{\arraystretch}{1.2}
% 更改了 tabular 定义以包含所有 |
\begin{tabular}{l c c c c}
\toprule
\multirow{2}{*}{Inputs} & \multicolumn{2}{c}{HyperFAS} & \multicolumn{2}{c}{FaceNet} \\ % 为 FaceNet 添加了 |
\cline{2-5} % 移除了 \cline 
& ACS $\uparrow$ & Acc $\uparrow$ & top-1 Acc $\uparrow$ & top-5 Acc $\uparrow$ \\
\hline
Clean & 0.952 & 70.70\% & 78.26\% & 87.68\% \\
\hline
+CMUA & 0.946 & 69.14\% & 77.55\% & 87.07\% \\
+FOUND & 0.948 & 70.31\% & 77.56\% & 87.07\% \\
+DWT & 0.948 & 70.10\% & 77.56\% & 87.07\% \\
+TSDF & 0.941 & 69.75\% & 77.55\% & 87.07\% \\
+AEF & \textbf{0.949} & \textbf{70.44\%} & \textbf{77.56\%} & \textbf{87.07\%} \\
\bottomrule
\end{tabular}
\end{table}
\begin{table*}[t]
\centering
\caption{Results of hold-out evaluations. Bold text indicates black-box models, whereas white-box models are denoted by *.}
\label{tab:hold_out_experiments}
\renewcommand{\arraystretch}{1.1}
\setlength{\tabcolsep}{7pt}
\begin{tabular}{lcccccccccc}
\toprule
\multirow{2}{*}{Model} & \multicolumn{5}{c}{$SR_{mask} \uparrow$} & \multicolumn{5}{c}{$L_{2mask} \uparrow$} \\
\cmidrule(lr){2-6} \cmidrule(lr){7-11}
& FOUND & CMUA & DWT & TSDF & AEF & FOUND & CMUA & DWT & TSDF & AEF \\
\midrule
% 第一组：HiSD 为黑盒
StarGAN* & 100.00\% & 100.00\% & 100.00\% & 100.00\% & 100.00\% & 0.28 & 0.24 & 0.20 & \textbf{0.54} & 0.20 \\
AGGAN* & 100.00\% & 99.85\% & 100.00\% & 100.00\% & 99.86\% & 0.23 & 0.22 & 0.18 & \textbf{0.25} & 0.24 \\
AttGAN* & 99.04\% & 78.20\% & 2.55\% & 94.33\% & \textbf{99.91\%} & 0.20 & 0.09 & 0.01 & 0.12 & \textbf{0.24} \\ 
\textbf{HiSD} & 2.69\% & 0.08\% & 0.05\% & 0.11\% & \textbf{3.63\%} & 0.03 & 0.00 & 0.00 & 0.02 & \textbf{0.04} \\
\midrule
% 第二组：AttGAN 为黑盒
StarGAN* & 100.00\% & 100.00\% & 100.00\% & 100.00\% & 100.00\% & 0.33 & 0.37 & 0.28 & 0.50 & \textbf{0.52} \\
AGGAN* & 100.00\% & 99.93\% & 100.00\% & 100.00\% & 100.00\% & 0.22 & 0.22 & 0.16 & \textbf{0.31} & 0.24 \\
\textbf{AttGAN} & 1.36\% & 0.03\% & 0.00\% & 0.04\% & \textbf{4.03\%} & 0.01 & 0.00 & 0.00 & 0.00 & \textbf{0.04} \\
HiSD* & 99.12\% & 99.65\% & 18.12\% & 99.31\% & \textbf{99.99\%} & 0.24 & 0.18 & 0.03 & 0.17 & \textbf{0.26} \\
\midrule
% 第三组：AGGAN 为黑盒 (发生黑盒迁移衰减)
StarGAN* & 100.00\% & 100.00\% & 100.00\% & 100.00\% & 100.00\% & 0.37 & 0.57 & 0.20 & \textbf{0.75} & 0.20 \\
\textbf{AGGAN} & 99.45\% & 99.20\% & 94.65\% & 99.77\% & \textbf{99.82\%} & 0.17 & 0.17 & 0.11 & 0.17 & \textbf{0.19} \\
AttGAN* & 98.89\% & 79.12\% & 2.23\% & 45.65\% & \textbf{99.26\%} & 0.19 & 0.10 & 0.01 & 0.05 & \textbf{0.21} \\
HiSD* & 90.41\% & 91.64\% & 16.88\% & 91.32\% & \textbf{93.35\%} & 0.11 & 0.12 & 0.03 & 0.11 & \textbf{0.15} \\
\midrule
% 第四组：StarGAN 为黑盒 (发生黑盒迁移衰减)
\textbf{StarGAN} & 100.00\% & 99.81\% & 95.12\% & 99.98\% & 100.00\% & 0.17 & 0.16 & 0.13 & 0.18 & \textbf{0.19} \\
AGGAN* & 100.00\% & 100.00\% & 100.00\% & 100.00\% & 100.00\% & 0.32 & 0.26 & 0.17 & \textbf{0.47} & 0.14 \\
AttGAN* & 98.79\% & 81.34\% & 1.95\% & 94.41\% & \textbf{99.88\%} & 0.20 & 0.10 & 0.01 & 0.13 & \textbf{0.23} \\
HiSD* & 96.77\% & 96.03\% & 17.20\% & 97.88\% & \textbf{98.07\%} & 0.17 & 0.16 & 0.03 & 0.17 & \textbf{0.19} \\
\bottomrule
\end{tabular}
\end{table*}
\begin{table}[htbp]
\centering
\caption{Quantitative evaluation of cross-architecture black-box transferability ($SR_{mask}$) driven by a single source model. The white-box source model is denoted by *.}
\label{tab:single_source_black_box}
\renewcommand{\arraystretch}{1.3}
\setlength{\tabcolsep}{8pt} 
\begin{tabular}{lcccc}
\toprule
\multirow{2}{*}{Method} & \multicolumn{4}{c}{$SR_{mask} \uparrow$} \\
\cmidrule(lr){2-5}
& StarGAN* & AGGAN & HiSD & AttGAN \\
\midrule
CMUA & 100.00\% & 99.74\% & 0.28\% & 0.00\% \\
FOUND & 100.00\% & 99.67\% & 1.32\% & 0.02\% \\
DWT & 100.00\% & 98.97\% & 0.04\% & 0.00\% \\
TSDF & 100.00\% & 99.85\% & 0.58\% & 0.00\% \\
\textbf{AEF} & 100.00\% & \textbf{99.99\%} & \textbf{2.91\%} & \textbf{0.08\%} \\
\bottomrule
\end{tabular}
\end{table}
\begin{table}[h!]
\centering
\caption{Ablation Study on the Efficacy of the Adaptive Equilibrium Component.}
\label{tab:ablation_adaptive_tuning_detailed}
\setlength{\tabcolsep}{4pt} % Adjust column spacing
\renewcommand{\arraystretch}{1.3} % Adjust row height
% 更改了 tabular 定义以包含 |
\begin{tabular}{l l c c c}
\toprule
Configuration & Model & SRmask $\uparrow$ & SSIM $\downarrow$ & L2mask $\uparrow$ \\
\hline
\multirow{5}{*}{\parbox{2cm}{\centering  Static Weighting}}
& StarGAN & 100.00\% & 0.26 & 0.23 \\
& AttGAN & 97.01\% & 0.64 & 0.15 \\
& HiSD & 92.10\% & 0.85 & 0.10 \\
& AGGAN & 100.00\% & 0.71 & \textbf{0.20} \\
\cline{2-5} % 替换 \cmidrule
& Average & 97.28\% & 0.62 & 0.17 \\
\hline
\multirow{5}{*}{\parbox{2cm}{\centering Adaptive Equilibrium }}
& StarGAN & \textbf{100.00\%} & \textbf{0.13} & \textbf{0.38} \\
& AttGAN & \textbf{99.65\%} & \textbf{0.55} & \textbf{0.21} \\
& HiSD & \textbf{99.85\%} & \textbf{0.74} & \textbf{0.17} \\
& AGGAN & \textbf{100.00\%} & \textbf{0.68} & 0.18 \\
\cline{2-5} % 替换 \cmidrule
& Average & \textbf{99.88\%} & \textbf{0.52} & \textbf{0.24} \\
\bottomrule
\end{tabular}
\end{table}
\begin{table}[h]
\centering
\caption{Effect of the feature shift hyperparameter $\alpha$ on interruption efficacy.}
\label{tab:ablation_alpha}
\setlength{\tabcolsep}{5pt}
\renewcommand{\arraystretch}{1.0}
\small 
\begin{tabular}{cccccc}
\toprule
\multicolumn{2}{c}{Configuration} & \multicolumn{4}{c}{Deepfake Model Performance} \\
\cmidrule(lr){1-2} \cmidrule(lr){3-6}
\textbf{$\alpha$} & Metric & HiSD & AttGAN & StarGAN & AGGAN \\
\midrule
\multirow{4}{*}{0.2} 
    & L2mask↑ & 0.17 & 0.20 & \textbf{0.43} & 0.20 \\
    & SRmask↑ & 99.50\% & \textbf{99.79\%} & 100.00\% & 100.00\% \\
    & PSNR↓ & 15.20 & 15.34 & 8.93 & 17.67 \\
    & SSIM↓ & 0.76 & 0.55 & 0.19 & 0.70 \\
\cmidrule(r){1-6}
\multirow{4}{*}{0.4} 
    & L2mask↑ & 0.15 & 0.21 & 0.39 & 0.19 \\
    & SRmask↑ & 99.32\% & 99.63\% & 100.00\% & 100.00\% \\
    & PSNR↓ & 16.21 & \textbf{15.15} & 9.35 & 18.24 \\
    & SSIM↓ & 0.76 & 0.55 & 0.17 & 0.70 \\
\cmidrule(r){1-6}
\multirow{4}{*}{0.6}
    & L2mask↑ & 0.16 & 0.21 & 0.38 & \textbf{0.20} \\
    & SRmask↑ & 99.75\% & 99.74\% & 100.00\% & 100.00\% \\
    & PSNR↓ & 15.09 & 15.29 & 9.03 & 17.45 \\
    & SSIM↓ & 0.75 & 0.55 & 0.15 & 0.68 \\
\cmidrule(r){1-6}
\multirow{4}{*}{0.8}
    & L2mask↑ & \textbf{0.17} & \textbf{0.21} & 0.38 & 0.18 \\
    & SRmask↑ & \textbf{99.85\%} & 99.65\% & \textbf{100.00\%} & \textbf{100.00\%} \\
    & PSNR↓ & 15.56 & 15.35 & 9.66 & \textbf{17.40} \\
    & SSIM↓ & \textbf{0.74} & \textbf{0.55} & \textbf{0.13} & \textbf{0.68} \\
\cmidrule(r){1-6}
\multirow{4}{*}{1.0}
    & L2mask↑ & 0.17 & 0.20 & 0.38 & 0.18 \\
    & SRmask↑ & 99.64\% & 99.70\% & 100.00\% & 100.00\% \\
    & PSNR↓ & \textbf{14.91} & 15.95 & \textbf{8.66} & 17.62 \\
    & SSIM↓ & 0.75 & 0.60 & 0.18 & 0.69 \\
\bottomrule
\end{tabular}
\end{table}
\begin{table}[htbp]
\centering
\caption{Ablation Study on the Temperature ($T$).}
\label{tab:temperature_analysis}
\renewcommand{\arraystretch}{1.3} 
\setlength{\tabcolsep}{1.5 pt} % 使用 mm 控制列距，更符合排版直觉
\begin{tabular}{ccccc}
\toprule
$T$ & Avg $SR_{mask} \uparrow$ & Std Dev (\%) $\downarrow$ & Avg $L_{2mask} \uparrow$ & Avg SSIM $\downarrow$ \\
\midrule
0.1 & \textbf{99.88\%} & \textbf{0.14} & 0.24 & \textbf{0.52} \\
0.5 & 98.15\% & 2.37 & 0.29 & 0.53 \\
1.0 & 97.80\% & 2.92 & 0.33 & 0.54 \\
2.0 & 97.61\% & 3.12 & 0.34 & 0.54 \\
3.0 & 95.50\% & 7.95 & \textbf{0.35} & 0.56 \\
\bottomrule
\end{tabular}
\end{table}

\subsection{Ablation Studies}
\subsubsection{Analysis of the Adaptive Equilibrium Component}
The objective is to demonstrate that this dynamic weighting mechanism is superior to a standard, non-adaptive ensemble approach for achieving balanced, cross-model generalization. In this experiment, we compare the full AEF method with a simple uniform averaging from all target models. The results presented in Table \ref{tab:ablation_adaptive_tuning_detailed} confirm the critical role of the adaptive strategy. The version of our framework using uniform weighting showed a marked decrease in overall interrupted performance. Although it performed adequately on certain models, its effectiveness is inconsistent, particularly against more robust architectures. In contrast, the full AEF framework with Adaptive Tuning maintained a consistently high success rate of interruption across the entire set of models. For instance, the adaptive approach achieved an average SRmask of 99.88\%, a significant improvement over 97.28\%, achieved by the uniform weighting baseline. This confirms that our adaptive mechanism successfully focuses the optimization on the most challenging models, thereby preventing the interruption from overfitting to easier targets and ensuring a universal and stronger perturbation. This result confirms that the Adaptive Tuning component is crucial to achieving the high level of generalization.

\subsubsection{Hyperparameter Analysis of Feature Shift}
The interruption efficacy of the feature enhancement module relies on the hyperparameter $\alpha$, which constrains the feature shift scale. Cross-model evaluation on four diverse architectures (Table \ref{tab:ablation_alpha}) reveals a non-monotonic relationship between interruption efficacy and $\alpha$. At a low setting ($\alpha = 0.2$), the perturbation fails to effectively disrupt deep network features. Optimal performance is achieved at $\alpha = 0.8$, where L2mask values for HiSD and AttGAN reach 0.17 and 0.21, respectively, and StarGAN's SSIM drops to 0.13. However, further increasing $\alpha$ to 1.0 exacerbates gradient conflicts during joint optimization, causing AttGAN's SSIM to rebound to 0.60 and degrading overall effectiveness. Balancing cross-architecture disruption limits and joint optimization stability, $\alpha = 0.8$ is established as the optimal configuration.
\subsubsection{Analysis of Temperature on Adaptive Equilibrium}
To determine an effective temperature and analyze its impact on adversarial equilibrium, the framework's performance is evaluated across a range of T values from 0.1 to 3.0. The results presented in Table \ref{tab:temperature_analysis} reveal a clear and consistent trend.

When the value of T is 0.1, the framework achieves a peak SRmask of 99.88\%. Crucially, this configuration also yields the lowest Standard Deviation of 0.14\%. This result demonstrates a near-perfect adversarial equilibrium across all models.

This analysis also highlights the superior efficiency of the T=0.1 setting. While the average output disruption or Avg L2mask at T=3.0 is numerically higher at 0.35 compared to 0.24 at T=0.1, setting T=3.0 is highly inefficient. It produces a larger but highly uneven average output disruption that fails to consistently surpass the success threshold, resulting in a low 95.50\% success rate and a complete loss of equilibrium. In contrast, the T=0.1 setting produces an extremely consistent output disruption. This strong equilibrium ensures nearly every attack surpasses the defined success threshold, achieving the 99.88\% peak success rate with a more efficient average output disruption of 0.24.

Increasing the temperature beyond 0.1 results in a clear performance decline in both effectiveness and equilibrium. Based on this comprehensive analysis, T=0.1 is identified as the optimal configuration, achieving maximal effectiveness and the strongest equilibrium with the highest efficiency.
\vspace{-1em}
\section{Conclusion}
In this paper, we propose the Adaptive Equilibrium Framework that solves the cross-architecture transferability defects in existing interruption perturbations. AEF addresses the interruption imbalance problem and weak destructive effect when faced with different deepfake models. The framework's efficacy derives from two synergistic components: adaptive equilibrium process to dynamically weight model losses, and deep feature enhancement process to disrupt essential intermediate representations. This dual mechanism ensures that the final perturbation is powerful and balanced, targeting common and specific vulnerabilities to achieve universal interruption. Comprehensive evaluation shows that AEF addresses the interruption imbalance problem between different architectures. Furthermore, the framework requires less training time to generate perturbations, thus reducing computational costs. The AEF guides the optimization process to an equilibrium state, ensuring consistent effectiveness against both resistant and susceptible deepfake models.

\printcredits

\section*{CRediT authorship contribution statement}

\textbf{Hongrui Zheng:} Conceptualization, Methodology, Software, Validation, Formal analysis, Investigation, Data curation, Writing -- original draft. \textbf{Liejun Wang:} Conceptualization, Resources, Supervision, Funding acquisition, Writing -- review \& editing. \textbf{Zhiqing Guo:} Methodology, Supervision, Project administration, Writing -- review \& editing.

\section*{Declaration of competing interest}
The authors declare that they have no known competing financial interests or personal relationships that could have appeared to influence the work reported in this paper.
\section*{Acknowledgments}
This work was supported in part by the National Natural Science Foundation of China under Grant 62462060, Grant 62302427, and Grant 62472368, in part by the Central Government Guides Local Science and Technology Development
Fund Projects under Grant ZYYD2026ZY21.
\section*{Data availability}
The datasets analyzed during the current study (CelebA, LFW, and FaceForensics++) are publicly available repositories and can be accessed via their respective official project websites. The source code and implementation details that support the findings of this study are available at \url{https://anonymous.4open.science/r/AEF-1259/}.
% Loading bibliography style file
\bibliographystyle{cas-model2-names}

\end{document}